\documentclass[runningheads]{llncs}
\usepackage[T1]{fontenc}
\usepackage{graphicx,verbatim,booktabs,multirow}
\begin{document}
\title{CUPA-T2*: Covariance-Aware Uncertainty Propagation and Alignment for T2* Mapping in Accelerated MRI}
\titlerunning{CUPA-T2*}
%

\author{Gideon N. L. Rouwendaal\inst{1,2}* \and
Natascha Niessen\inst{2,3} \and
Hannah Eichhorn\inst{1,2} \and
Dirk H. J. Poot\inst{4} \and
Christine Preibisch\inst{5,6} \and
Julia A. Schnabel\inst{1,2,6,7,8}}
\authorrunning{G. Rouwendaal et al.}
%
\institute{
Institute of Machine Learning in Biomedical Imaging, Helmholtz Munich, Germany \and
School of Computation, Information \& Technology, Technical University of Munich, Germany \and
GE HealthCare, Munich, Germany \and
Department of Radiology and Nuclear Medicine, Erasmus MC, Rotterdam, The Netherlands \and
School of Medicine \& Health, Technical University of Munich, Germany \and
Department of Neuroradiology, Klinikum rechts der Isar, Technical University of Munich, Germany \and
Munich Center for Machine Learning (MCML), Germany \and
School of Biomedical Engineering \& Imaging Sciences, King's College London, UK 
\\
\email{*gideon.rouwendaal@tum.de}}
  
\maketitle              
\begin{abstract}
Quantitative T2* maps have strong potential for biomarker discovery but are limited by long scan times, rendering them impractical in clinical settings. Significant acceleration can be achieved through undersampling in k-space combined with learning-based reconstruction. However, reconstruction artifacts and noise can propagate into downstream T2* fitting, degrading its accuracy. We introduce CUPA-T2*, a framework that explicitly propagates voxel-wise inter-echo uncertainty from stochastic Monte Carlo dropout reconstructions to downstream T2* fitting via covariance-aware sampling. T2* fitting is performed with a heteroscedastic MLP and a correlation-based regularizer that encourages alignment between predicted variance and reconstruction uncertainty. Experiments on accelerated brain MRI data show tissue-dependent behavior: CUPA-T2* achieves competitive overall T2* fitting performance and improves white-matter performance at higher accelerations. Compared with a heteroscedastic baseline, the proposed framework substantially increases alignment between reconstruction uncertainty and predicted T2* variance, while also revealing a trade-off with calibration (ECE) and selective prediction performance (AURC). CUPA-T2* enables reconstruction uncertainty-aware T2* fitting and delivers voxel-wise uncertainty maps to support the interpretation of quantitative T2* estimates. Our code is available at https://github.com/compai-lab/2026-miccai-rouwendaal.


\end{abstract}

\section{Introduction}
Quantitative MRI (qMRI) estimates physical tissue parameters from multi-contrast acquisitions, improving comparability across scanners and facilitating biomarker discovery. In brain imaging, T2* mapping enables the assessment of iron deposition, which is associated with neurodegenerative pathologies such as Parkinson’s disease, Alzheimer’s disease, and multiple sclerosis (particularly in rim-positive white matter (WM) lesions) \cite{haacke2005imaging,zhou2025imaging,pyatigorskaya2020iron,gillen2018ironmicroglia}. However, qMRI techniques rely on multi-contrast acquisitions, which substantially increase scan time. Acceleration via k-space undersampling is therefore essential and has significantly contributed to the broader adoption of qMRI. This also renders the reconstruction problem ill-posed, leading to artifacts and noise that can propagate into downstream parameter estimation.

Deep learning methods have substantially advanced accelerated qMRI reconstruction from undersampled data \cite{heckel2024deep}. In practice, qMRI typically involves a two-stage pipeline consisting of multi-contrast image reconstruction followed by voxel-wise parameter estimation. While these approaches substantially improve reconstruction quality and parameter prediction accuracy, principled uncertainty quantification across reconstruction and downstream parameter estimation remains limited.

\begin{figure}[b]
\includegraphics[width=\textwidth]{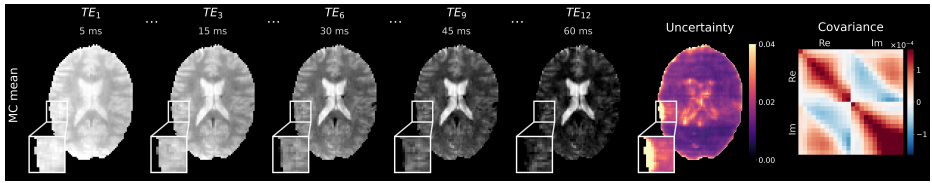}
\caption{Mean reconstructions across echo times (TE), computed over MC-dropout samples, with the corresponding voxel-wise reconstruction uncertainty map (summarized as the RMS marginal standard deviation) and an example voxel-wise inter-echo covariance matrix.}
\label{fig:TE_RMS_Dcay}
\end{figure}

Predictive uncertainty in deep learning can be decomposed into aleatoric (data-inherent) and epistemic (model) components \cite{kendall2017uncertainties,gal2016dropout,tanno2019uncertainty}. In accelerated MRI, undersampling yields non-unique reconstructions, with multiple solutions consistent with the same measured k-space. In stochastic learning-based models, this ambiguity appears as epistemic uncertainty in the reconstructed complex signal. In multi-echo T2* mapping, joint reconstruction of echoes or shared sampling patterns can introduce correlations between echoes into the uncertainty estimates. Figure \ref{fig:TE_RMS_Dcay} illustrates this effect via a representative voxel-wise inter-echo covariance matrix with pronounced off-diagonal structure. Ignoring inter-echo correlations may degrade the robustness of T2* parameter estimates, leading to potential misdiagnosis.

Most existing (q)MRI approaches either neglect reconstruction uncertainty or summarize it using voxel-wise scalar variances \cite{sun2025guiding,edupuganti2021uncertainty,ekanayake2025pixcue,tanno2019uncertainty,glang2020deepcest}, ignoring the structured covariance induced by undersampling. Uncertainty propagation has been explored in accelerated MRI pipelines \cite{fischer2023uncertainty,feiner2023propagation}. In qMRI, Sun et al. \cite{sun2025guiding} introduced phase-wise uncertainty guidance within a two-stage reconstruction–fitting framework using Monte Carlo (MC) dropout. Nevertheless, explicit modeling of covariance and enforcement of cross-stage alignment between reconstruction-derived epistemic and downstream predicted aleatoric uncertainty remain unaddressed.

In this work, we present \textbf{C}ovariance-Aware \textbf{U}ncertainty \textbf{P}ropagation and \textbf{A}lignment for robust \textbf{T2*} mapping (CUPA-T2*), a two-stage framework that explicitly preserves and propagates inter-echo covariance into a heteroscedastic T2* regression model while enforcing structural alignment between reconstruction epistemic uncertainty and predicted aleatoric uncertainty. To the best of our knowledge, this is the first work to combine covariance-aware uncertainty propagation with an explicit objective that aligns reconstruction-derived uncertainty with predicted T2* variance. The main contributions can be summarized as follows:

\begin{enumerate}
    \item We introduce covariance-aware sampling that transfers structured reconstruction uncertainty into heteroscedastic T2* regression.
    
    \item We propose a Root Mean Square (RMS) based correlation regularizer that promotes structural alignment between reconstruction-derived epistemic uncertainty and predicted aleatoric uncertainty.
\end{enumerate}

\section{Methods}
\begin{figure}[t]
\includegraphics[width=\textwidth]{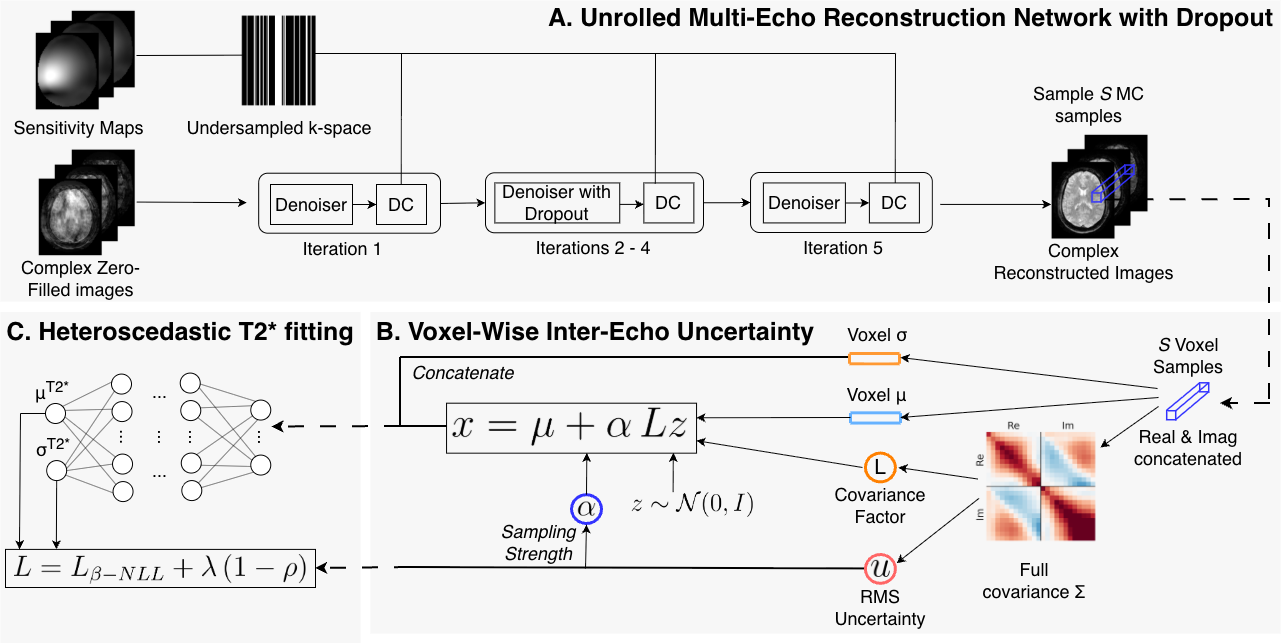}
\caption{Overview of the proposed CUPA-T2* framework. (A) An unrolled multi-echo reconstruction network with MC dropout. (B) Voxel-wise inter-echo covariance is estimated from MC samples and factorized to enable covariance-aware sampling, modulated by RMS uncertainty. (C) A heteroscedastic T2* regressor predicts mean and variance, trained with $\beta$-NLL and RMS-based cross-stage uncertainty losses.}
\label{fig:schematic_overview}
\end{figure}

Figure \ref{fig:schematic_overview} shows the proposed CUPA-T2* framework, comprising three stages: (A) An unrolled reconstruction network with MC dropout \cite{eichhorn2026motionrobust,gal2016dropout}, (B) voxel-wise inter-echo covariance estimation with RMS summarization, and (C) heteroscedastic T2* regression with covariance-aware sampling and uncertainty alignment. The stages are described in detail below.

\subsection{Unrolled multi-echo reconstruction network with dropout}
We adopt the unrolled multi-echo reconstruction architecture described in \cite{eichhorn2026motionrobust}, which is validated on the public \textit{T2*-MOVE} dataset \cite{eichhorn2026t2move}, also used in this study. The network reconstructs complex-valued multi-echo images from undersampled multi-coil k-space via $N=5$ iterations alternating CNN denoising and data consistency (DC) steps. To enable stochastic sampling, dropout ($p=0.3$) is inserted into intermediate denoiser layers and kept active during inference \cite{gal2016dropout}. After training, a set of $S=100$ stochastic reconstructions ${x^{(s)}}_{s=1}^{S}$ per slice is generated, serving as the basis for voxel-wise inter-echo uncertainty estimation.

\subsection{Voxel-wise inter-echo uncertainty and sampling}
\label{sec:voxelwise_uncertainty}
\begin{figure}[b]
\includegraphics[width=\textwidth]{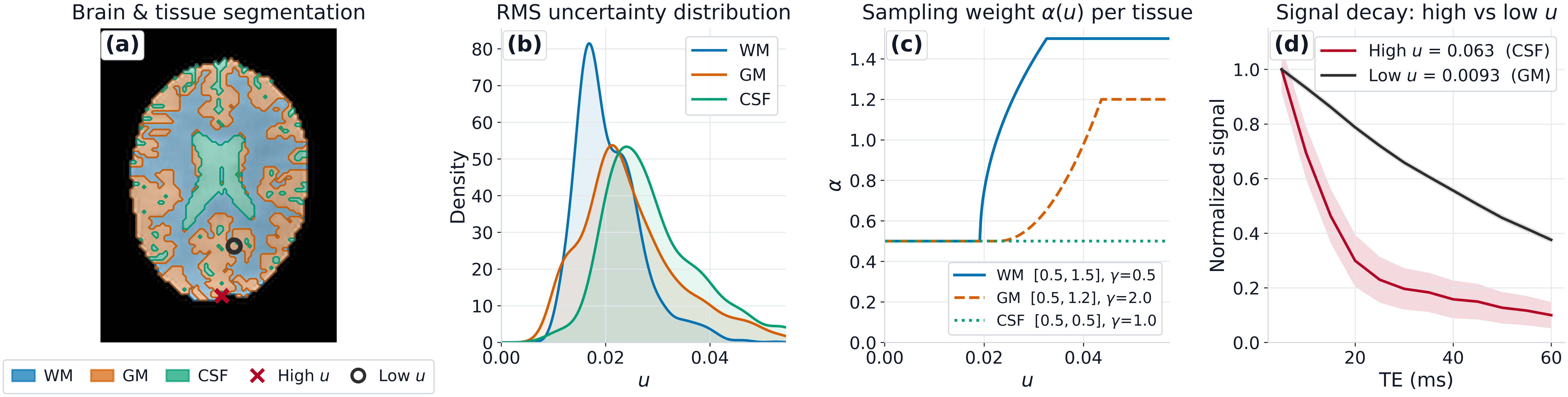}
\caption{Tissue-dependent uncertainty and sampling. (a) Tissue segmentation with example high- and low-$u$ voxels. (b) Uncertainty distributions per tissue. (c) Sampling schedules $\alpha(u)$. (d) Mean $\pm$ 1 SD MC signal decay for high- and low-$u$ voxels.}
\label{fig:RMS_dist}
\end{figure}

From the $S$ MC reconstructions with $E=12$ echoes, each voxel $v$ is represented as a real vector $x^{(s)}_{v} \in \bbbr^{2E}$, formed by concatenating the real and imaginary components across echoes, from which the mean $\mu_v$ and covariance $\Sigma_v$ are computed. To propagate the reconstruction uncertainty to the T2* fitting stage, a covariance factor $L_v$ is calculated such that $\Sigma_v \approx L_v L_v^\top$, where $L_v\in \bbbr^{2E\times 2E}$ is obtained via Cholesky decomposition. This yields a full-rank factor that preserves the estimated inter-echo covariance structure. We additionally derive a scalar summary for alignment and noise-schedule control by computing the RMS marginal standard deviation, given as $u_v=\sqrt{\frac{1}{2E}\,\mathrm{tr}(\Sigma_v)}$. Given $\mu_v$ and factor $L_v$, we generate uncertainty-aware samples as 
\begin{equation} 
x^{(s)}_{v} = \mu_v + \alpha_v\, L_v z^{(s)}
\end{equation}
where $z^{(s)} \sim \mathcal{N}(0, I)$ and $\alpha_v$ controls stochastic noise proportional to voxel-wise uncertainty. The sampling coefficient is defined as $\alpha_v = \alpha_{\min} + (\alpha_{\max} - \alpha_{\min}) \hat{u}_v^{\gamma}$, where $\gamma$ determines how strongly sampling strength increases with uncertainty. Figure \ref{fig:RMS_dist} illustrates tissue-dependent uncertainty, motivating tissue-aware $\alpha$ scheduling. The resulting schedule adjusts the noise injection level based on uncertainty, keeping it low in clean voxels and higher in noisy ones. The magnitude $\tilde{x}_v$ of the resulting complex-valued samples $x_v^s$ serves as input to the parameter fitting network.

\subsection{Covariance-aware heteroscedastic T2* fitting and predictive uncertainty}
\label{sec:heteroscedastic_regression}
We employ a heteroscedastic multi-layer perceptron (MLP) that jointly predicts the voxel-wise conditional mean and aleatoric variance of T2*, given the sampled representation $\tilde{x}_v$. 
Specifically, the MLP outputs a mean $\hat{\mu}_{v}^{\mathrm{T2*}}$ and a log-variance  $\log(\sigma_{v}^{\mathrm{T2*}}) =  \hat{\ell}_{v}^{\mathrm{T2*}}$. The heteroscedastic MLP includes dropout layers that remain active during inference to enable MC-based epistemic uncertainty estimation \cite{tanno2019uncertainty}. Training is performed using the $\beta$-negative log-likelihood ($\beta$-NLL) loss \cite{seitzer2022pitfalls},

\begin{equation} 
L_{\beta\mathrm{-NLL}}(x_v)
=
\Bigl[ \exp(\hat{\ell}_{v}^{\mathrm{T2*}}) \Bigr]^{\beta}_{\mathrm{sg}}
\frac{1}{2}
\left(
\exp\!\big(-\hat{\ell}_{v}^{\mathrm{T2*}}\big)
\big(y_{v}^{\mathrm{T2*}}-\hat{\mu}_{v}^{\mathrm{T2*}}\big)^2
+
\hat{\ell}_{v}^{\mathrm{T2*}}
\right)
\end{equation}
where $[\cdot]_{\mathrm{sg}}$ denotes the stop-gradient operation. We set $\beta=0.5$ following \cite{seitzer2022pitfalls}.

Reference T2* targets ($y_v^{T2*}$) are obtained from fully sampled k-space reconstructions by voxel-wise non-linear least-squares fitting of a mono-exponential decay model across all 12 echoes within a brain mask and T2* values clipped at 200\,ms.

Although reconstruction uncertainty arises from epistemic variability in the stochastic reconstruction network, it induces irreducible stochastic variability in the regression inputs and thus acts as effective input-dependent aleatoric variability during parameter estimation. To promote structural alignment between this reconstruction-derived uncertainty and the predicted conditional variance, we introduce an RMS-based correlation regularizer. Specifically, we align the predicted log-variance with the reconstruction-derived RMS uncertainty $u_v$ by adding $\rho = \mathrm{Pearson}(\hat{\ell}_{v}^{\mathrm{T2*}}, \log u_v)$ to $L = L_{\beta\mbox{-}NLL} + \lambda (1-\rho)$, where $\lambda$ controls alignment strength. Importantly, the regularizer promotes correspondence between the two uncertainty patterns rather than assuming a formal equivalence between upstream epistemic and downstream aleatoric uncertainty. Because RMS uncertainty exhibits tissue-dependent distributions, a global correlation would be dominated by high-variance tissues. We therefore compute the Pearson correlation separately for WM, gray matter (GM), and cerebrospinal fluid (CSF) and average them with equal weights. Although stratification narrows the within-tissue dynamic range, it prevents variance-driven bias and promotes balanced structural alignment across tissues.

At inference, predictive uncertainty is decomposed into aleatoric and epistemic components \cite{kendall2017uncertainties,tanno2019uncertainty}. 
Using $M=100$ MC-dropout forward passes, aleatoric uncertainty is estimated as the mean predicted variance $\frac{1}{M}\sum_{m=1}^{M} \exp(\hat{\ell}_v^{(m)})$, and epistemic uncertainty as the variance of the predictive means $\mathrm{Var}_m(\hat{\mu}_v^{(m)})$. The total predictive uncertainty is their sum.

The MLP has five fully connected layers (hidden dimension 64), trained with Adam (learning rate 0.001, batch size 1024), using early stopping (patience 40) on mean WM and GM NRMSE. Following \cite{sun2025guiding}, per-echo standard deviations from $S$ MC-dropout reconstructions in the magnitude domain are concatenated to the magnitude signal, yielding a $\bbbr^{2E}$ feature vector. Inputs are normalized by the first echo. The RMS-alignment weight is set to $\lambda=0.05$, as it achieved the best validation performance across acceleration rates in most settings. During validation and testing, the MC mean and standard deviation are used.

\section{Experiments and Results}\label{sec:experiments}
\subsection{Dataset}
We use the publicly available \textit{T2*-MOVE} dataset for quantitative T2* mapping of the human brain \cite{eichhorn2026t2move}. Fully sampled k-space data are retrospectively undersampled using a single Cartesian random mask along the phase-encoding (PE) direction, simulating acceleration factors $R = 2, 3, 4$. We focus on this range because $R=4$ already reduces scan time to 55 seconds, while higher accelerations leave too few PE lines for reliable T2* fitting. The 22 subjects are split subject-wise into 14/3/4 for training, validation, and testing, yielding 381/87/104 slices after discarding slices containing less than 20\% brain tissue. 

\subsection{Evaluation}
We compare CUPA-T2* to a non-heteroscedastic MLP (PUQ) \cite{sun2025guiding} and a heteroscedastic baseline (Het) \cite{kendall2017uncertainties}, without covariance-aware sampling and RMS alignment.  
T2* map quality is assessed using NRMSE and SSIM within each brain mask and per tissue. Aleatoric uncertainty performance is assessed using the calibration error (ECE) computed from empirical coverage at $1\sigma$, $2\sigma$, and $3\sigma$, area under the risk-coverage curve (AURC), and Pearson correlation ($\rho$) between reconstruction RMS and predicted aleatoric variance. Wilcoxon signed-rank tests are employed for statistical testing, using $p=0.001$.

\subsection{Results and Discussion}
Table \ref{tab:exp14_overall_metrics} shows the NRMSE and SSIM on the test set for the different models and acceleration rates. All methods show consistently good performance, with only small differences between them. Overall, PUQ achieves the lowest global NRMSE and the highest SSIM. However, a tissue-specific analysis (see Table~\ref{tab:exp14_tissue_nrmse}) reveals that CUPA-T2* reduces WM error at acceleration $R = 3$ and $R = 4$. A possible explanation is that white matter is relatively homogeneous and exhibits fewer physiological fluctuations, yielding more reliable inter-echo covariance estimates and, therefore, less-noisy propagated samples. In contrast, higher uncertainty in CSF under stronger undersampling may limit its benefit. This suggests that covariance-aware propagation is most beneficial in regions with coherent interecho structure, but less effective in high-uncertainty regions.

\begin{figure}[t]
\centering
\includegraphics[width=0.72\textwidth]{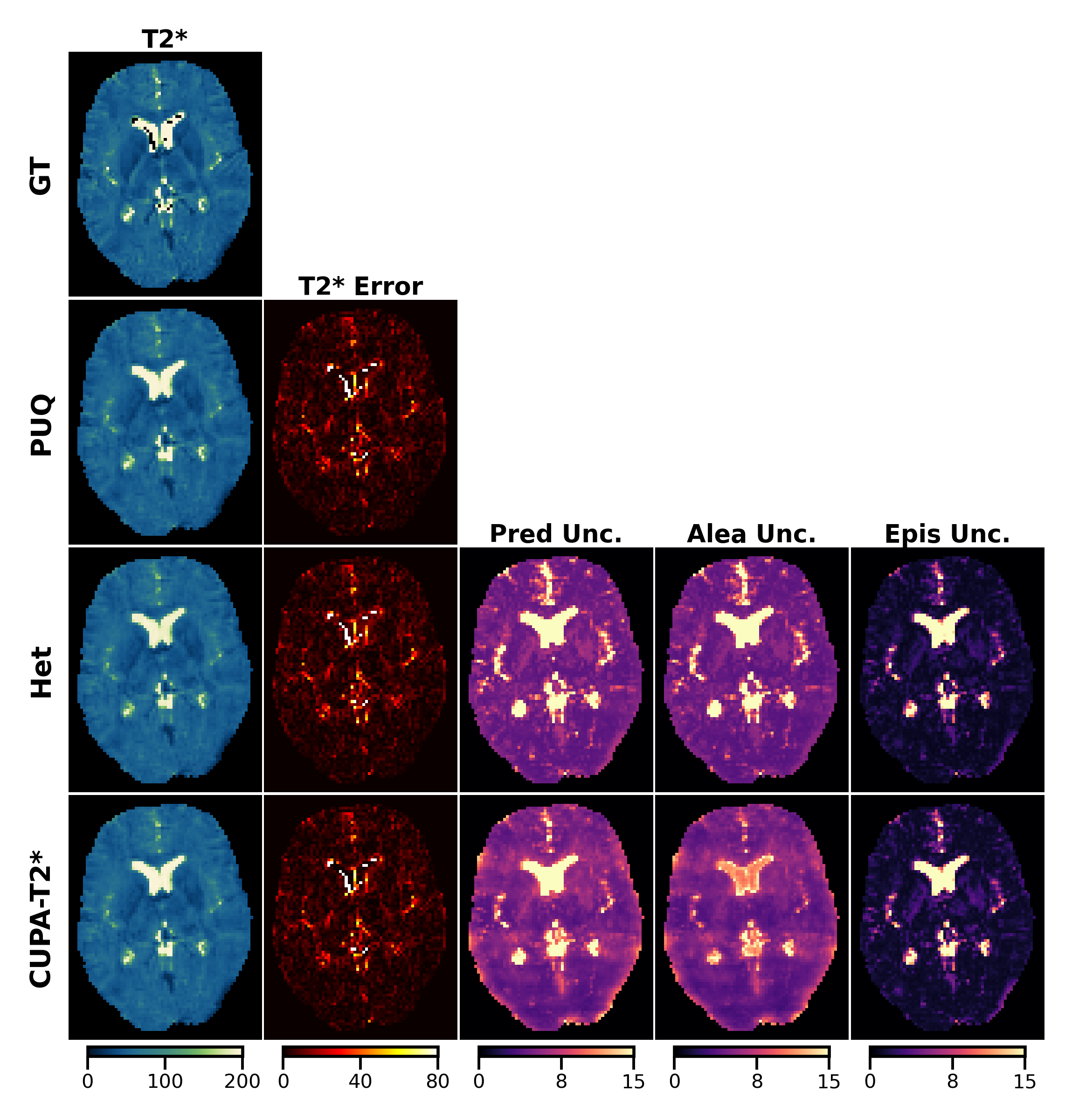}
\caption{Qualitative comparison of T2* estimation and uncertainty. First column: reference and predicted T2* maps. Second column: absolute error maps. Columns 3–5: total predictive uncertainty, aleatoric uncertainty, and epistemic uncertainty, respectively.}
\label{fig:QualiResults}
\end{figure}


\begin{table}[!t]
\centering
\caption{T2* fitting performance (NRMSE, SSIM) over all tissues across acceleration rates. Best metrics in \textbf{bold} and $^*$ indicating statistically superior to all others.}
\label{tab:exp14_overall_metrics}
{\fontsize{8}{9}\selectfont
\setlength{\tabcolsep}{5pt}
\renewcommand{\arraystretch}{1.15}
\begin{tabular}{l cc cc cc}
\toprule
\multirow{2}{*}{Method}
  & \multicolumn{2}{c}{2$\times$}
  & \multicolumn{2}{c}{3$\times$}
  & \multicolumn{2}{c}{4$\times$} \\
\cmidrule(lr){2-3}\cmidrule(lr){4-5}\cmidrule(lr){6-7}
  & NRMSE $\downarrow$ & SSIM $\uparrow$
  & NRMSE $\downarrow$ & SSIM $\uparrow$
  & NRMSE $\downarrow$ & SSIM $\uparrow$ \\
\midrule
PUQ      & \textbf{0.125}$^{*}$ & \textbf{0.964}$^{*}$ & \textbf{0.181}$^{*}$ & \textbf{0.928}$^{*}$ & \textbf{0.235}$^{*}$ & \textbf{0.893}$^{*}$ \\
Het      & 0.128 & 0.962 & 0.183 & 0.926 & 0.236 & 0.891 \\
CUPA-T2* & 0.127 & 0.963 & 0.188 & 0.921 & 0.242 & 0.885 \\
\bottomrule
\end{tabular}
}
\end{table}


\begin{table}
\centering
\caption{Tissue-specific T2* NRMSE under different acceleration rates. Best metrics in \textbf{bold} and $^*$ indicating statistically superior to all others. CUPA-T2* achieves the lowest WM NRMSE at $R=3,4$.}
\label{tab:exp14_tissue_nrmse}
{\fontsize{8}{9}\selectfont
\setlength{\tabcolsep}{3.5pt}
\renewcommand{\arraystretch}{1.15}
\begin{tabular}{l ccc ccc ccc}
\toprule
\multirow{2}{*}{Method}
  & \multicolumn{3}{c}{2$\times$}
  & \multicolumn{3}{c}{3$\times$}
  & \multicolumn{3}{c}{4$\times$} \\
\cmidrule(lr){2-4}\cmidrule(lr){5-7}\cmidrule(lr){8-10}
  & WM $\downarrow$ & GM $\downarrow$ & CSF $\downarrow$
  & WM $\downarrow$ & GM $\downarrow$ & CSF $\downarrow$
  & WM $\downarrow$ & GM $\downarrow$ & CSF $\downarrow$ \\
\midrule
PUQ      & 0.100 & \textbf{0.117}$^{*}$ & \textbf{0.166}$^{*}$ & 0.137 & \textbf{0.168} & \textbf{0.250}$^{*}$ & 0.174 & \textbf{0.209} & \textbf{0.334}$^{*}$ \\
Het      & 0.100 & 0.118 & 0.174 & 0.138 & 0.169 & 0.256 & 0.173 & 0.211 & 0.337 \\
CUPA-T2* & \textbf{0.099} & 0.119 & 0.171 & \textbf{0.135}$^{*}$ & 0.171 & 0.270 & \textbf{0.168}$^{*}$ & 0.210 & 0.359 \\
\bottomrule
\end{tabular}
}
\end{table}

\subsubsection{Uncertainty.} Table \ref{tab:exp14_unc_metrics_overall} summarizes the uncertainty metrics. While Het achieved the lowest AURC and ECE, it failed to capture the correlation between reconstruction uncertainty and predicted variance. In contrast, CUPA-T2* substantially increases this relationship. Qualitatively, Figure \ref{fig:QualiResults} confirms this correlation. CUPA-T2* uncertainty maps highlight regions more prone to reconstruction artifacts, such as voxel boundaries and brain-background interfaces. At the same time, AURC and ECE increase, indicating a trade-off between cross-stage uncertainty alignment and downstream uncertainty quality. A possible reason is that the current objective aligns predicted T2* variance with epistemic reconstruction uncertainty, which reflects only part of the total upstream predictive
uncertainty.


\begin{table}[t]
\centering
\caption{Overall uncertainty metrics across acceleration rates. Best metrics in \textbf{bold} and $^*$ indicating statistically superior to all others.}
\label{tab:exp14_unc_metrics_overall}
{\fontsize{8}{9}\selectfont
\setlength{\tabcolsep}{2pt}
\renewcommand{\arraystretch}{1.15}
\begin{tabular}{l ccc ccc ccc}
\toprule
\multirow{2}{*}{Method}
  & \multicolumn{3}{c}{2$\times$}
  & \multicolumn{3}{c}{3$\times$}
  & \multicolumn{3}{c}{4$\times$} \\
\cmidrule(lr){2-4}\cmidrule(lr){5-7}\cmidrule(lr){8-10}
  & AURC $\downarrow$ & $\rho$ $\uparrow$ & ECE $\downarrow$
  & AURC $\downarrow$ & $\rho$ $\uparrow$ & ECE $\downarrow$
  & AURC $\downarrow$ & $\rho$ $\uparrow$ & ECE $\downarrow$ \\
\midrule
Het      & \textbf{3.398}$^{*}$ & 0.280 & \textbf{0.031}$^{*}$ & \textbf{4.560}$^{*}$ & 0.259 & \textbf{0.033}$^{*}$ & \textbf{5.637}$^{*}$ & 0.189 & \textbf{0.036}$^{*}$ \\
CUPA-T2* & 3.471 & \textbf{0.766}$^{*}$ & 0.046 & 4.814 & \textbf{0.728}$^{*}$ & 0.051 & 5.910 & \textbf{0.651}$^{*}$ & 0.047 \\
\bottomrule
\end{tabular}
}
\end{table}

\begin{table}[!t]
\centering
\caption{Ablation: tissue-specific $T_2^*$ NRMSE and overall uncertainty at $R=4$. Best metrics in \textbf{bold} and $^*$ indicating statistically superior to all others.}
\label{tab:exp14_tissue_nrmse_plus_overall_unc}
{\fontsize{8}{9}\selectfont
\setlength{\tabcolsep}{4pt}
\renewcommand{\arraystretch}{1.15}
\begin{tabular}{l ccc ccc}
\toprule
\multirow{2}{*}{Method}
  & \multicolumn{3}{c}{Quantitative (NRMSE)}
  & \multicolumn{3}{c}{Overall uncertainty} \\
\cmidrule(lr){2-4}\cmidrule(lr){5-7}
  & WM $\downarrow$ & GM $\downarrow$ & CSF $\downarrow$
  & AURC $\downarrow$ & $\rho$ $\uparrow$ & ECE $\downarrow$ \\
\midrule
CUPA-T2*       & 0.168 & 0.210 & 0.359 & 5.910 & 0.651 & 0.047 \\
$\alpha = 0$   & 0.173 & 0.211 & \textbf{0.335}$^{*}$ & 5.902 & 0.641 & 0.042 \\
$\Sigma I$     & 0.173 & 0.213 & 0.340 & 5.901 & 0.628 & 0.049 \\
$\lambda = 0.0$  & \textbf{0.167}$^{*}$ & 0.211 & 0.386 & 5.772 & 0.250 & 0.060 \\
$\lambda = 0.01$ & 0.169 & \textbf{0.210} & 0.354 & \textbf{5.723} & 0.370 & \textbf{0.042} \\
$\lambda = 0.1$  & \textbf{0.167}$^{*}$ & 0.212 & 0.360 & 6.319 & \textbf{0.839}$^{*}$ & 0.072 \\
\bottomrule
\end{tabular}
}
\end{table}

\subsubsection{Ablation Study (R = 4).}
We evaluated the impact of individual CUPA-T2* components by ablating: (i) covariance-aware sampling ($\alpha=0$), (ii) inter-echo covariance by replacing it with the diagonal covariance ($\Sigma I$), and (iii) alignment strength ($\lambda$). Table \ref{tab:exp14_tissue_nrmse_plus_overall_unc} summarizes the T2* fitting and uncertainty performance at R = 4. Disabling covariance-aware sampling ($\alpha=0, \Sigma I$) degrades WM NRMSE, but improves CSF, confirming benefits for low-uncertainty tissues. Removing the regularizer ($\lambda=0$) yields the largest drop in $\rho$. Introducing a small alignment correlation ($\lambda=0.01$) increases $\rho$ while achieving the best AURC and ECE. Further increasing to $\lambda=0.1$ maximizes $\rho$, but worsens AURC and ECE, supporting the trade-off between correlation and AURC and ECE.

\section{Conclusion}
In this work, we proposed CUPA-T2*, a framework that moves beyond scalar-only uncertainty by propagating inter-echo covariance from stochastic reconstructions into heteroscedastic fitting. This enables voxel-wise T2* uncertainty maps, informed by structured reconstruction uncertainty through explicit alignment. Our results indicate tissue-dependent behavior: CUPA-T2* reduces fitting errors in WM, whereas performance in high-uncertainty CSF is less robust. Overall, CUPA-T2* significantly improves correlation between reconstruction-derived epistemic uncertainty and predicted aleatoric variance. The proposed framework can naturally extend to other quantitative parameters that exhibit inter‑contrast correlations. Future work should explore aligning with total predictive uncertainty rather than only epistemic uncertainty.

\begin{credits}
\subsubsection{\ackname} G.N.L.R. is funded by the European Union under the MSCA-DN project IQ-BRAIN. Views and opinions expressed are, however, those of the author(s) only and do not necessarily reflect those of the European Union or the European Research Executive Agency. Neither the European Union nor the granting authority can be held responsible for them. C.P. is supported by the Deutsche Forschungsgemeinschaft 
(DFG, German Research Foundation) under project number 563563598. D.H.J. Poot receives research funding from GE Healthcare. This work is supported by the DAAD programme Konrad Zuse Schools of Excellence in Artificial Intelligence sponsored by the German Federal Ministry of Research, Technology and Space.
\subsubsection{\discintname}
NN is an employee of GE HealthCare, Munich. All other authors declare that they do not have any financial or non-financial conflict of interests.
\end{credits}

%
%
%

\begin{thebibliography}{8}
\bibitem{haacke2005imaging}
Haacke, E.M., Cheng, N.Y.C., House, M.J., Liu, Q., Neelavalli, J.,
Ogg, R.J., Khan, A., Ayaz, M., Kirsch, W., Obenaus, A.: Imaging iron
stores in the brain using magnetic resonance imaging.
Magnetic Resonance Imaging \textbf{23}(1), 1--25 (2005)

\bibitem{kendall2017uncertainties}
Kendall, A., Gal, Y.: What uncertainties do we need in Bayesian deep learning for computer vision?
Advances in Neural Information Processing Systems \textbf{30} (2017)

\bibitem{zhou2025imaging}
Zhou, C.-H., Zhu, Y.-C.:
Imaging of cerebral iron as an emerging marker for brain aging, neurodegeneration, and cerebrovascular diseases.
Brain Sciences \textbf{15}(9), 944 (2025).
\doi{10.3390/brainsci15090944}

\bibitem{eichhorn2026t2move}
Eichhorn, H. et al.: T2*-MOVE: An Open k-Space Dataset to Address Motion
in T2* Quantification from Gradient Echo MRI.
Nefeli, Helmholtz Zentrum M{\"u}nchen (2026).
\url{https://nefeli.helmholtz-munich.de/records/9hqjb-tfw77}

\bibitem{eichhorn2026motionrobust}
Eichhorn, H., Spieker, V., Hammernik, K., Saks, E., Felsner, L.,
Weiss, K., Preibisch, C., Schnabel, J.A.:
Motion-robust T$_2^*$ quantification from low-resolution gradient echo brain MRI
with physics-informed deep learning.
Magnetic Resonance in Medicine \textbf{95}, 346--362 (2026).
\doi{10.1002/mrm.70050}

\bibitem{gal2016dropout}
Gal, Y., Ghahramani, Z.: 
Dropout as a Bayesian approximation: Representing model uncertainty in deep learning. 
In: Proceedings of the 33rd International Conference on Machine Learning (ICML), 
PMLR 48, 1050--1059 (2016)

\bibitem{sun2025guiding}
Sun, H., Li, Z., Du, C., Li, H., Wang, Y., Chen, H.:
Guiding quantitative MRI reconstruction with phase-wise uncertainty.
In: International Conference on Medical Image Computing and Computer-Assisted Intervention (MICCAI), 
pp.~248--257. Springer (2025)


\bibitem{kingma2015adam}
D.~P. Kingma and J.~Ba,
\newblock Adam: A Method for Stochastic Optimization,
\newblock in \emph{Proceedings of the International Conference on Learning Representations (ICLR)}, 2015.


\bibitem{wang2004ssim}
Z.~Wang, A.~C. Bovik, H.~R. Sheikh, and E.~P. Simoncelli,
\newblock Image quality assessment: From error visibility to structural similarity,
\newblock \emph{IEEE Transactions on Image Processing}, vol.~13, no.~4, pp.~600--612, 2004.
\newblock doi:10.1109/TIP.2003.819861.

\bibitem{seitzer2022pitfalls}
Seitzer, M., Tavakoli, A., Antic, D., Martius, G.:
On the Pitfalls of Heteroscedastic Uncertainty Estimation with Probabilistic Neural Networks.
arXiv preprint arXiv:2203.09168 (2022).
\doi{10.48550/arXiv.2203.09168}

\bibitem{pyatigorskaya2020iron}
Pyatigorskaya, N., Sanz-Morère, C.B., Gaurav, R., Biondetti, E., 
Valabregue, R., Santin, M., Yahia-Cherif, L., Lehéricy, S.:
Iron imaging as a diagnostic tool for Parkinson's disease: 
A systematic review and meta-analysis.
Frontiers in Neurology \textbf{11}, 366 (2020).
\doi{10.3389/fneur.2020.00366}


\bibitem{shafieizargar2023systematic}
Shafieizargar, B., Byanju, R., Sijbers, J., Klein, S., 
den Dekker, A.J., Poot, D.H.J.:
Systematic review of reconstruction techniques for accelerated quantitative MRI.
Magnetic Resonance in Medicine \textbf{90}(3), 1172--1208 (2023).
\doi{10.1002/mrm.29721}

\bibitem{edupuganti2021uncertainty}
Edupuganti, V., Mardani, M., Vasanawala, S., Pauly, J.:
Uncertainty quantification in deep MRI reconstruction.
IEEE Transactions on Medical Imaging \textbf{40}(1), 239--250 (2021).
\doi{10.1109/TMI.2020.3025065}

\bibitem{ekanayake2025pixcue}
Ekanayake, M., Pawar, K., Chen, Z., Egan, G., Chen, Z.:
PixCUE: Joint uncertainty estimation and image reconstruction in MRI using deep pixel classification.
Journal of Imaging Informatics in Medicine \textbf{38}(4), 2071--2084 (2025).

\bibitem{tanno2019uncertainty}
Tanno, R., Worrall, D., Kaden, E., Ghosh, A., Grussu, F., 
Bizzi, A., Sotiropoulos, S.N., Criminisi, A., Alexander, D.C.:
Uncertainty quantification in deep learning for safer neuroimage enhancement.
arXiv preprint arXiv:1907.13418 (2019).
\doi{10.48550/arXiv.1907.13418}

\bibitem{glang2020deepcest}
Glang, F., Deshmane, A., Prokudin, S., et al.:
DeepCEST 3T: Robust MRI parameter determination and uncertainty quantification with neural networks—application to CEST imaging of the human brain at 3T
Magnetic Resonance in Medicine \textbf{84}, 450--466 (2020).
\doi{10.1002/mrm.28117}

\bibitem{heckel2024deep}
Heckel, R., Jacob, M., Chaudhari, A., Perlman, O., Shimron, E.:
Deep learning for accelerated and robust MRI reconstruction.
MAGMA \textbf{37}(3), 335--368 (2024).
\doi{10.1007/s10334-024-01173-8}


\bibitem{Jeelani2020AMT}
Jeelani, H., Yang, Y., Zhou, R., Kramer, C.M., Salerno, M., Weller, D.S.:
A myocardial T1-mapping framework with recurrent and U-Net convolutional neural networks.
In: 2020 IEEE 17th International Symposium on Biomedical Imaging (ISBI),
pp.~1941--1944 (2020).
\doi{10.1109/ISBI45749.2020.9098653}


\bibitem{liu2019mantis}
Liu, F., Feng, L., Kijowski, R.:
MANTIS: Model-augmented neural network with incoherent k-space sampling for efficient MR parameter mapping.
Magnetic Resonance in Medicine \textbf{82}, 174--188 (2019).
\doi{10.1002/mrm.27707}

\bibitem{jun2021dopamine}
Jun, Y., Shin, H., Eo, T., Kim, T., Hwang, D.:
Deep model-based magnetic resonance parameter mapping network (DOPAMINE) for fast T1 mapping using variable flip angle method.
Medical Image Analysis \textbf{70}, 102017 (2021).
\doi{10.1016/j.media.2021.102017}


\bibitem{zimmermann2024pinqi}
Zimmermann, F.F., Kolbitsch, C., Schuenke, P., Kofler, A.:
PINQI: An end-to-end physics-informed approach to learned quantitative MRI reconstruction.
IEEE Transactions on Computational Imaging \textbf{10}, 628--639 (2024).
\doi{10.1109/TCI.2024.3388869}

\bibitem{fischer2023uncertainty}
Fischer, P., Thomas, K., Baumgartner, C.F.:
Uncertainty estimation and propagation in accelerated MRI reconstruction.
In: International Workshop on Uncertainty for Safe Utilization of Machine Learning in Medical Imaging,
pp.~84--94. Springer, Cham (2023).

\bibitem{feiner2023propagation}
Feiner, L.F., Menten, M.J., Hammernik, K., Hager, P.,
Huang, W., Rueckert, D., Braren, R.F., Kaissis, G.:
Propagation and attribution of uncertainty in medical imaging pipelines.
In: International Workshop on Uncertainty for Safe Utilization of Machine Learning in Medical Imaging,
pp.~1--11. Springer, Cham (2023).

\bibitem{gillen2018ironmicroglia}
Gillen, K.M., Mubarak, M., Nguyen, T.D., Pitt, D.:
Significance and In Vivo Detection of Iron-Laden Microglia in White Matter Multiple Sclerosis Lesions.
Frontiers in Immunology \textbf{9}, 255 (2018).
\doi{10.3389/fimmu.2018.00255}

\end{thebibliography}
%

\end{document}